\documentclass[sigconf]{acmart}
\usepackage{amsmath}
\usepackage{subfigure}

\AtBeginDocument{%
  \providecommand\BibTeX{{%
    \normalfont B\kern-0.5em{\scshape i\kern-0.25em b}\kern-0.8em\TeX}}}

\copyrightyear{2021} 
\acmYear{2021} 
\setcopyright{acmcopyright}\acmConference[GLSVLSI '21]{Proceedings of the Great Lakes Symposium on VLSI 2021}{June 22--25, 2021}{Virtual Event, USA}
\acmBooktitle{Proceedings of the Great Lakes Symposium on VLSI 2021 (GLSVLSI '21), June 22--25, 2021, Virtual Event, USA}
\acmPrice{15.00}
\acmDOI{10.1145/3453688.3461755}
\acmISBN{978-1-4503-8393-6/21/06}

\begin{document}

%%
%% The "title" command has an optional parameter,
%% allowing the author to define a "short title" to be used in page headers.
\title{On the Adversarial Robustness of Quantized Neural Networks}

%%
%% The "author" command and its associated commands are used to define
%% the authors and their affiliations.
%% Of note is the shared affiliation of the first two authors, and the
%% "authornote" and "authornotemark" commands
%% used to denote shared contribution to the research.
\author{Micah Gorsline, James Smith, and Cory Merkel}
%\authornote{Both authors contributed equally to this research.}
\email{{mdg2741,jss1612,cemeec}@rit.edu}
%\orcid{1234-5678-9012}
%\authornotemark[1]
\affiliation{%
  \institution{Brain Lab\\Rochester Institute of Technology}
  \streetaddress{1 Lomb Memorial Drive}
  \city{Rochester}
  \state{NY}
  \country{USA}
  \postcode{14623}
}

%%
%% By default, the full list of authors will be used in the page
%% headers. Often, this list is too long, and will overlap
%% other information printed in the page headers. This command allows
%% the author to define a more concise list
%% of authors' names for this purpose.
%\renewcommand{\shortauthors}{Gorsline, Smith and Merkel}

%%
%% The abstract is a short summary of the work to be presented in the
%% article.
\begin{abstract}
Reducing the size of neural network models is a critical step in moving AI from a cloud-centric to an edge-centric (i.e. on-device) compute paradigm.  This shift from cloud to edge is motivated by a number of factors including reduced latency, improved security, and higher flexibility of AI algorithms across several application domains (e.g. transportation, healthcare, defense, etc.).  However, it is currently unclear how model compression techniques may affect the robustness of AI algorithms against adversarial attacks.  This paper explores the effect of quantization, one of the most common compression techniques, on the adversarial robustness of neural networks.  Specifically, we investigate and model the accuracy of quantized neural networks on adversarially-perturbed images.  Results indicate that for simple gradient-based attacks, quantization can either improve or degrade adversarial robustness depending on the attack strength.
\end{abstract}

%%
%% The code below is generated by the tool at http://dl.acm.org/ccs.cfm.
%% Please copy and paste the code instead of the example below.
%%
\begin{CCSXML}
<ccs2012>
<concept>
<concept_id>10010583.10010633.10010640.10010641</concept_id>
<concept_desc>Hardware~Application specific integrated circuits</concept_desc>
<concept_significance>500</concept_significance>
</concept>
<concept>
<concept_id>10010520.10010521.10010542.10010294</concept_id>
<concept_desc>Computer systems organization~Neural networks</concept_desc>
<concept_significance>500</concept_significance>
</concept>
<concept>
<concept_id>10010147.10010257.10010293.10010294</concept_id>
<concept_desc>Computing methodologies~Neural networks</concept_desc>
<concept_significance>500</concept_significance>
</concept>
</ccs2012>
\end{CCSXML}

\ccsdesc[500]{Hardware~Application specific integrated circuits}
\ccsdesc[500]{Computer systems organization~Neural networks}
\ccsdesc[500]{Computing methodologies~Neural networks}
%%
%% Keywords. The author(s) should pick words that accurately describe
%% the work being presented. Separate the keywords with commas.
\keywords{Adversarial machine learning, neural networks, quantization}

%%
%% This command processes the author and affiliation and title
%% information and builds the first part of the formatted document.
\maketitle

\section{Introduction}

State-of-the-art artificial neural networks (ANN) such as deep convolutional neural networks (CNN) contain hundreds of millions of parameters or more.  Despite their impressive results on tasks like image classification, modern ANNs' large size makes them very difficult to deploy on size, weight, and power (SWaP)-constrained hardware such as neuromorphic chips.  As a consequence, the promise of neuromorphic hardware, exemplified by chips like IBM TrueNorth \cite{merolla2014million} and Intel Loihi \cite{davies2018loihi}, is currently limited by the size and complexity of today's ANN models.  There are some ongoing efforts to address this issue by compressing neural network models through quantization/pruning and sharing techniques, low-rank factorization methods, transferred/compact weight methods, and knowledge distillation \cite{cheng2017survey}.  

Quantization, in particular, is a popular compression method due to its simplicity.  Quantized neural networks use reduced precision for weights, activations, and gradients to limit the computation and storage requirements.  A number of quantization strategies have been proposed \cite{zhou2017incremental,wang2019haq,dong2019hawq,courbariaux2016binarized,banner2019post}, which can generally be placed into two categories:  post-training quantization and quantized training.  Post-training quantization uses full precision during the training process and then performs post processing to reduce the precision of the trained parameters.  After the parameter precision has been reduced, the model can be deployed on neuromorphic hardware.  Quantized training, on the other hand, assesses the effects of parameter quantization during the training process, which usually leads to better results.  In addition, performing quantization iteratively during the training process is an attractive approach for neuromorphic systems because it enables on-chip training.  This is desirable since it improves model accuracy when faults are present (device variations, defects, noise, etc.) \cite{Merkel2014vlsid,alibart2013pattern} and enables fast continual learning.

An important side effect of model quantization is that it affects the margins of the ANN's decision boundaries, which directly influences how susceptible the model is to adversarial examples:  slightly perturbed inputs that cause high-confidence misclassifications.  Some previous works have begun to study the effects of quantization on ANNs' robustness to adversarial examples. Galloway et al. \cite{galloway2017attacking} show that neural networks with both weights and activations quantized to $\pm1$ provide some defense against iterative gradient-based attacks.  Bernhard et al. \cite{bernhardadversarial} argue that for appropriate attack strengths, quantization does not offer robust protection against adversarial attacks.  They also note, however, that there is a reduction in the transferability of attacks between models with different quantization levels.  In contrast, Duncan et al. \cite{duncan2020relative} show that quantization can in fact provide adversarial robustness relative to full-precision models.  They also confirm that transferability of adversarial examples between quantized models is reduced compared to full-precision models.  In \cite{lin2019defensive}, authors propose a defensive quantization technique that limits the amplification of input perturbations caused by the high gain of quantized activation functions, improving the robustness of quantized ANNs.  In addition, Song et al. \cite{song2020improving} have proposed an ANN re-training process to improve adversarial robustness of ANNs with quantized weights.    

%Despite all of their differences, neuromorphic system designs usually have one common characteristics:  They target small SWaP, which significantly limits the size and precision of the neural networks that can be configured on them.  Today's state-of-the-art deep neural networks like CNNs typically employ single or double precision-parameters (e.g. weights and biases) and computations (e.g. dot products and activation function evaluation), which significantly complicates or even precludes their direct deployment on neuromorphic hardware.  For example, a typical network size of $\sim$ 100 million parameters requires several gigabytes of storage, which is beyond the memory capacity of several SWaP-constrained edge computing platforms.  Therefore, there is a fundamental disagreement between the algorithm-level requirements of state-of-the-art neural networks and the hardware-level characteristics of neuromorphic systems. 

Here, we aim to shed new light on the robustness of quantized ANNs by building on previous work and providing the following contributions:
\begin{itemize}
\item A geometric model of the robustness of quantized ANNs against gradient-based attacks
\item Introduction of the concept of \textit{critical attack strength} that quantifies the attack strength at which quantization will not affect the ANN's accuracy
\item Comparison of adversarial robustness in quantized ANNs with different activation functions and different input dimensions
\end{itemize}

The rest of this paper proceeds as follows:  Section \ref{sec:background} provides background on adversarial attacks.  Section \ref{sec:model} describes the proposed geometric model of adversarial robustness.  In Section \ref{sec:results}, we show simulation results for ANN accuracy across different levels of weight quantization.  Section \ref{sec:conclusions} concludes this work.

\section{Background}
\label{sec:background}

Recent research focus in the area of adversarial machine learning (AML) is causing the deep learning community to question the safety, security, and trustworthiness of today's top-performing AI models.  AML concerns both the offensive and defensive measures associated with ML performance, security, and privacy.  This work focuses on the robustness of ML models to AML  evasion attacks (offense) \cite{joseph2018adversarial,vorobeychik2018adversarial,biggio2018wild}.  Evasion attacks exploit small-margin decision boundaries, where legitimate inputs to the model are perturbed just enough to move to them to a different decision region in the input space.  In the mid-2000's, evasion attacks were introduced as small perturbations to the content of emails, causing them to be misclassified by linear spam filters \cite{dalvi2004adversarial,lowd2005adversarial,lowd2005good}.  In 2014, Szegedy et al. \cite{szegedy2013intriguing} showed that imperceptible perturbations in the pixel space of images led to high-confidence misclassifications by CNNs.  Since then, the body of research related to AML, primarily evasion attacks and defenses, has grown rapidly in the domain of deep learning \cite{biggio2018wild}.  The goal of an evasion attack can be expressed as an optimization problem, where, for some model $\Pi$, a correctly-classified input $\mathbf{u}$, usually from the test or training set, is perturbed by $\mathbf{r}^{*}$ to maximize a loss function $\mathcal{L}_{\Pi}$ and cause $\Pi$'s classification of $\mathbf{u}^{\prime}=\mathbf{u}+\mathbf{r}^{*}$ to be different from $\mathbf{u}$'s ground truth label:

\begin{equation}
\begin{aligned}
\mathbf{r}^{*}= \underset{\mathbf{r}\in\mathcal{R},}{\arg\max}&\quad \mathcal{L}_{\Pi}(\mathbf{u}+\mathbf{r},l)\\
\textrm{s.t.}&\quad l^{\prime}\ne l
\label{eqn:evasion}
\end{aligned}
\end{equation}

\noindent where $l$ is $\mathbf{u}$'s ground truth label, $l^{\prime}$ is the model's label for $\mathbf{u}^{\prime}$, and $\mathcal{R}$ is the set of set of allowed perturbations.  $\mathbf{u}^{\prime}$ is called an adversarial example.  Often, the allowed set of perturbations takes the form of an $\ell^{p}$-norm constraint: $\mathcal{R}=\{\mathbf{r}\in\mathbb{R}^{N}:\:||\mathbf{r}||_{p}\le D\}$ where $N$ is the dimension of the input space. The $\ell^{p}$-norm of $\mathbf{r}$ is usually bounded in a way that the difference between  $\mathbf{u}$ and $\mathbf{u}^{\prime}$ is difficult or impossible to perceive by a human.  Attacks are usually performed using $\ell^{p}$-norms with $p=2$ or $p=\infty$.  However, $p=0$ and $p=1$ are also common.  $\mathcal{R}$ may also be formed using multiple $\ell^{p}$-norms, box constraints (e.g. bounding all inputs between a minimum and maximum value), or by choosing $\mathbf{r}$ as some type of transformation that imposes a dependence between the elements of $\mathbf{r}$ (e.g. affine transformations such as rotation, scaling, etc.).  A number of evasion attacks have been proposed based on (\ref{eqn:evasion}), differing primarily in the way they define $\mathcal{R}$, how much information they assume is known about $\Pi$ (whitebox vs. blackbox attacks), the way they approach the optimization procedure, and whether they are targeted (e.g. classifying a school bus image as an ostrich) or untargeted (e.g. classifying a school bus as anything other than a school bus).  In this paper, we focus on the untargeted version of the fast gradient sign method (FGSM) attack, which is a simple gradient-based evasion attack introduced by Goodfellow et al. \cite{fgsm}.  FGSM can be written as
\begin{equation}
\mathbf{r}^{*}=\epsilon\cdot\mathrm{sgn}\left(\nabla_\mathbf{u}\mathcal{L}_{\Pi}(\mathbf{u},l) \right)
\end{equation}
\noindent
where $\mathrm{sgn}(\cdot)$ is the sign function.  This attack is easy to apply in a white box setting when full details of $\Pi$ (i.e. structure and parameters) are known.

\section{Model Development}
\label{sec:model}

\begin{figure}
    \centering
    \subfigure[]{
    \includegraphics[width=0.45\columnwidth]{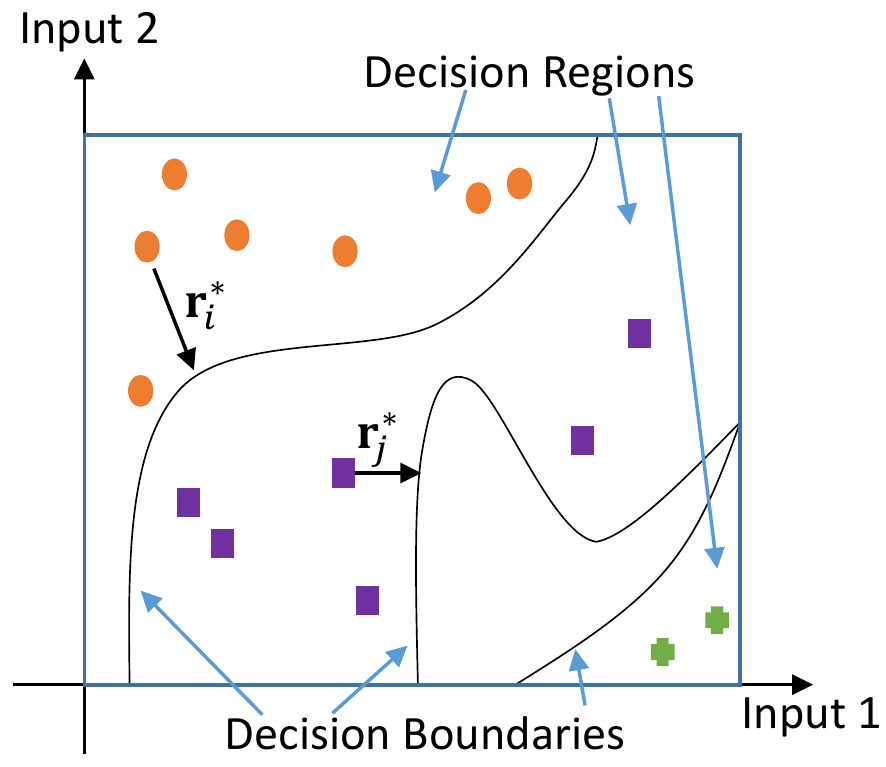}
    }
    \subfigure[]{
        \includegraphics[width=0.45\columnwidth]{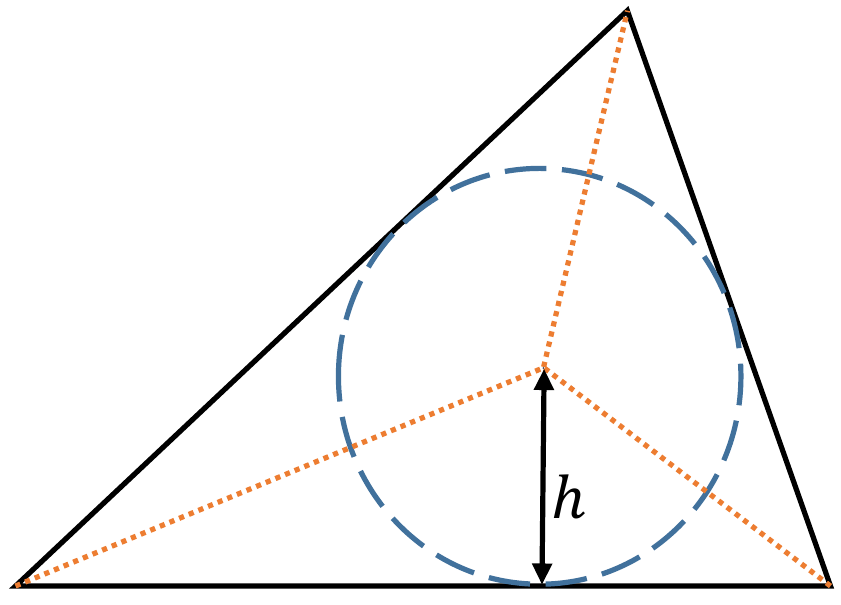}
    }
    \caption{Illustration of decision regions in a 2D input space.  (a) General non-linear decision boundaries dividing 3 classes in the input space.  (b) 2-simplex decision region divided into 3 simplices of equal height.}
    \label{fig:decisionregions}
\end{figure}

\begin{figure*}[!t]
\centering
\subfigure[]{
\includegraphics[width=0.3\textwidth]{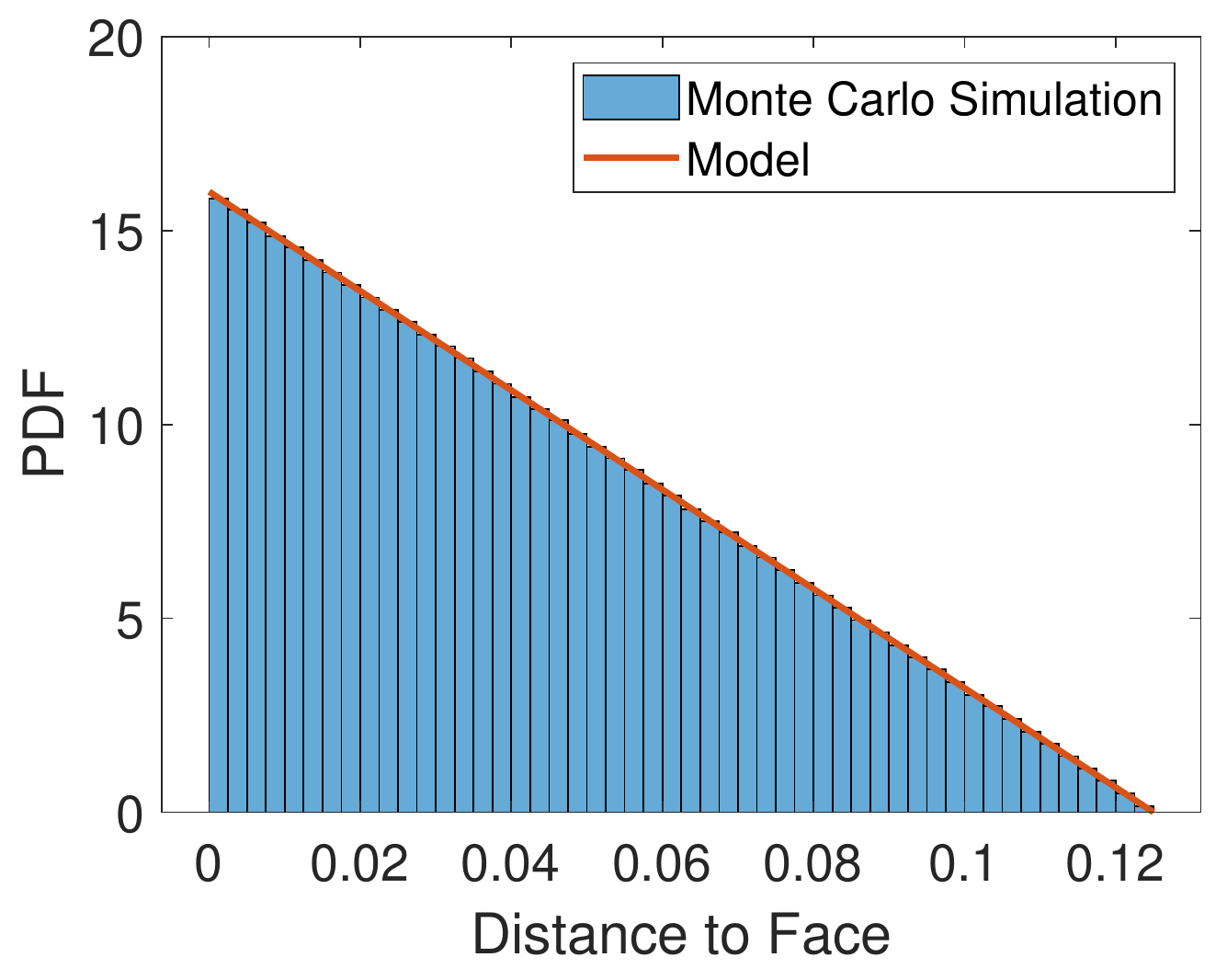}
}
\subfigure[]{
\includegraphics[width=0.3\textwidth]{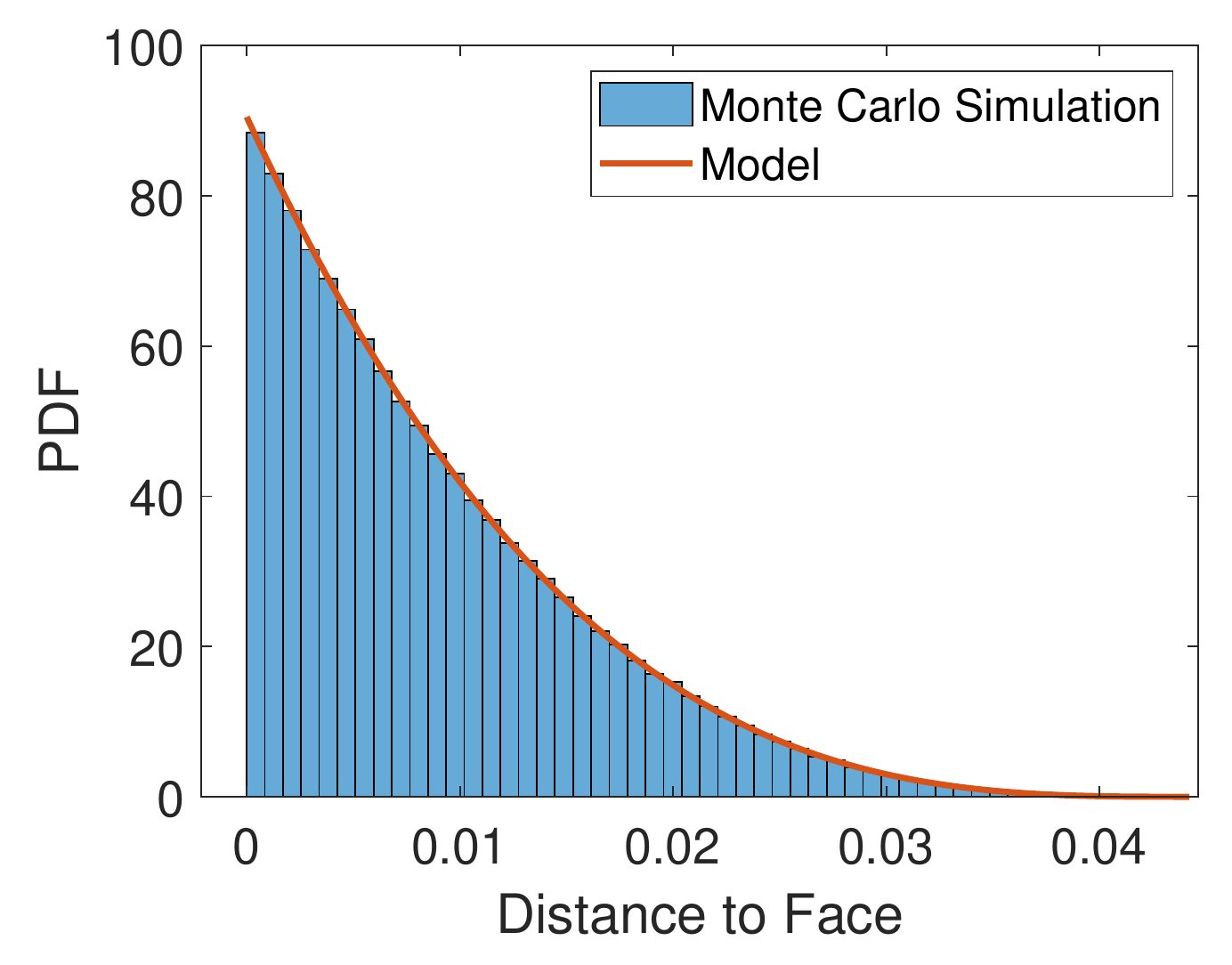}
}
\subfigure[]{
\includegraphics[width=0.3\textwidth]{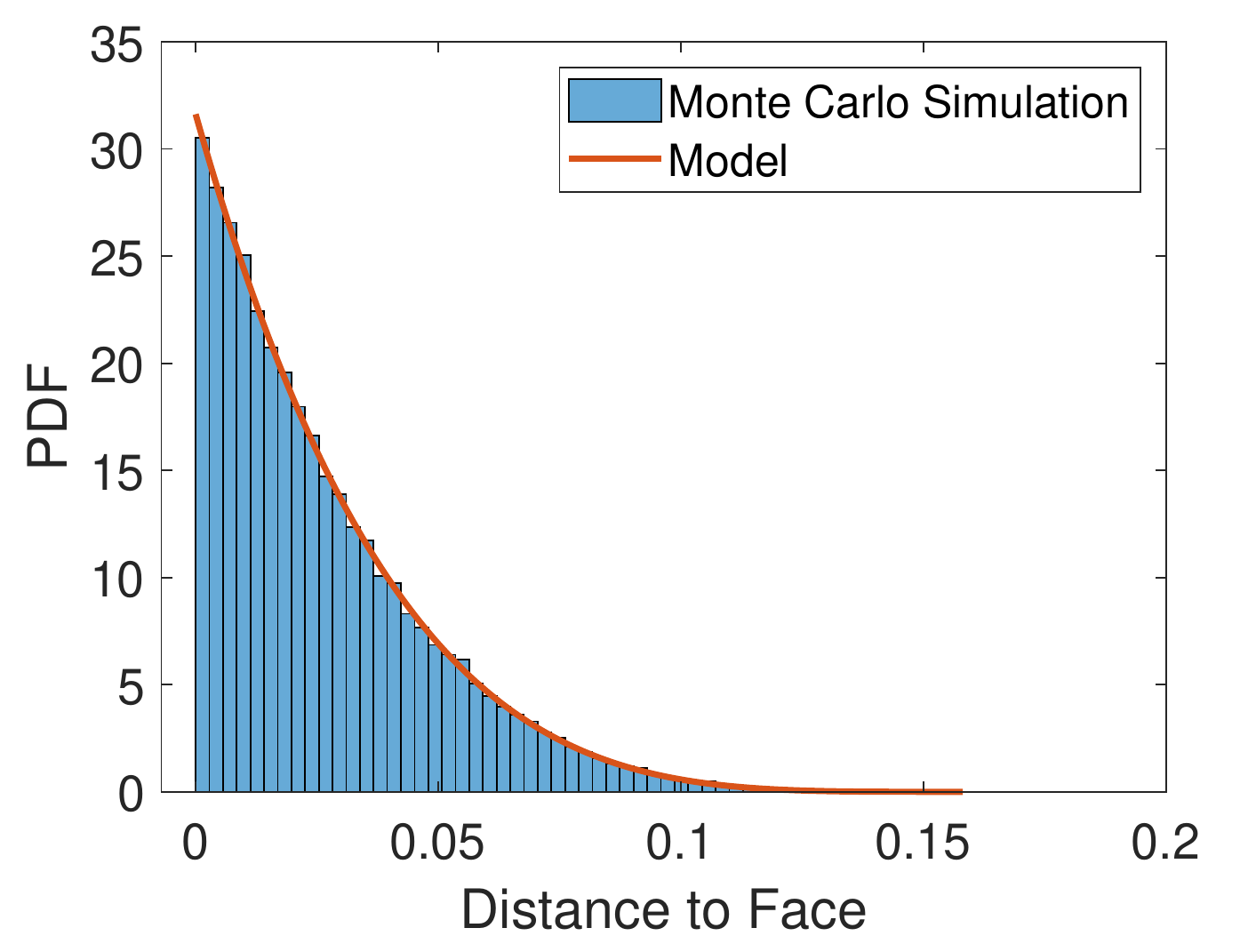}
}
\caption{Monte Carlo simulations of $f(\epsilon)$ for (a) $q=2$ and $n=2$, (b) $q=3$ and $n=4$, (c) $q=1$ and $n=5$.}
\label{fig:montecarlo}
\end{figure*}

The performance of ML models on adversarially-perturbed inputs, such as those derived from an FGSM attack, can be modeled by considering the distribution of distances between datapoints and decision boundaries.  Figure \ref{fig:decisionregions}(a) shows an example of a 2D input space with a number of decision boundaries which collectively define a number of decision regions.  The proximity of datapoints in each region to the closest boundary determines the robustness of the model against evasion attacks.  In this work, we assume that the input space is an $n$-dimensional hypercube.  In the simplest case, the decision regions will be defined by $n$-simplices.  Since here we are interested in $\ell^{2}$-bounded attacks, our goal is to determine the euclidean distance between datapoints and the faces of those simplices.  This will determine how quickly the accuracy of the network will degrade as points are moved in the direction of their loss gradient.  Figure \ref{fig:decisionregions}(b) shows our approach.  First, we find the incenter of the simplex and its radius given by
\begin{equation}
h=n\frac{V}{S}
\end{equation}
where $V$ and $S$ are the volume and surface area of the simplex, respectively.  Now, the simplex can be divided into $n+1$ smaller simplices of equal height $h$.  Every point inside one of the smaller simplices will be closest to the same face of the larger simplex.  Initially, we will assume that the points we are interested in (i.e. points from the training or test set) are uniformly distributed inside the simplex.  Therefore, the average $\ell^{2}$ distance $\bar{\epsilon}$ of a point to a face will be given by the component of the center of mass normal to the face:
\begin{equation}
\bar{\epsilon}=\frac{h}{n+1}=\frac{n}{n+1}\frac{V}{S}\approx\frac{V}{S}
\end{equation}
This simple expression is actually quite intuitive and captures concepts like the fact that smooth decision boundaries tend to reduce overfitting/improve adversarial robustness.  To find the volume-to-surface area ratio requires some assumptions about how the learning algorithms divides the input space.  Here, we use Draghici’s \cite{draghici2002capabilities} result that a hypercube of length $l$ can be shattered by a set of simplices $\{\mathcal{S}_i\}$ such that $sd(\mathcal{S}_{i})\le\mathrm{max}(l,l\sqrt{n}/2)$, where $sd(\mathcal{S}_{i})$ is the length of the longest edge of simplex $\mathcal{S}_{i}$.  Let $p=2^{q-1}$, where $q$ is the number of weight bits, and let $l=1/p$. Then $sd(\mathcal{S}_{i} )\le\mathrm{max}(2^{1-q},\sqrt{n}2^{-q})$.  For large $n$: $sd(\mathcal{S}_{i})\le\sqrt{n}2^{-q}$.  Now, we assume our space is divided by regular simplices with edge length $\sqrt{n}2^{-q}$.  The volume of a regular $n$-simplex with side length $s$ is \cite{rabinowitz1989volume}
\begin{equation}
V=\frac{s^{n}}{n!}\sqrt{\frac{n+1}{2^{n}}}.
\end{equation}
The surface area is just $n+1$ times the volume of one face:
\begin{equation}
S=\frac{(n+1)s^{n-1}}{(n-1)!}\sqrt{\frac{n}{2^{n-1}}}
\end{equation}
This results in:
\begin{equation}
\bar{\epsilon}\approx\frac{V}{S}=\frac{s}{(n+1)n}\sqrt{\frac{n+1}{2n}}=\frac{s}{\sqrt{2n^{3}(n+1)}}\approx\frac{s}{n^{2}\sqrt{2}}
\end{equation}
Plugging in $s=\sqrt{n}2^{-q}$:
\begin{equation}
\bar{\epsilon}\approx\frac{1}{2^{q+1/2}n^{3/2}}
\label{eqn:meandistfinal}
\end{equation}
From (\ref{eqn:meandistfinal}) and the assumptions made so far about the decision region shape and distribution of datapoints, we we see that the mean distance to the decision boundary decreases as the weight precision and the input dimensionality increase.  Therefore, we expect that there will be some adversarial protection provided by quantization.

We can also derive a probability distribution function $f$ for $\epsilon$ by finding a function that satisfies the following constraints
\begin{equation}
\bar{\epsilon}=\frac{h}{n+1}=\int\limits_{0}^{h}\epsilon f(\epsilon)\mathrm{d}\epsilon
\end{equation}
\begin{equation}
\int\limits_{0}^{h}f(\epsilon)\mathrm{d}\epsilon=1
\end{equation}
\begin{equation}
f(h)=0
\end{equation}
as well as the idea that $f$ should be monotonically decreasing from 0 to $h$.  The function that satisfies all of these constraints is
\begin{equation}
f(\epsilon)=\frac{n}{h^{n}}(h-\epsilon)^{n-1}
\label{eqn:pdf}
\end{equation}
Figure \ref{fig:montecarlo} shows Monte Carlo simulations of the distance to the decision boundary for a few different values of $q$ and $n$, along with the model in (\ref{eqn:pdf}).  We see excellent agreement between the model and simulation data.  The cumulative distribution function can be derived from the probability distribution function as
\begin{equation}
F(\epsilon)=1-\frac{1}{h^{n}}(h-\epsilon)^{n}
\end{equation}
Now, the accuracy degradation caused by an $\ell^{2}$-bounded attack can be modeled as
\begin{equation}
A_{r}\ge1-F(\epsilon)
\label{eqn:ra}
\end{equation}
where $A_{r}$ is the relative accuracy, defined as the ratio of the adversarial and clean accuracies (accuracy on unperturbed inputs).

\begin{figure*}[!th]
\centering
\subfigure[]{
\includegraphics[width=0.3\textwidth]{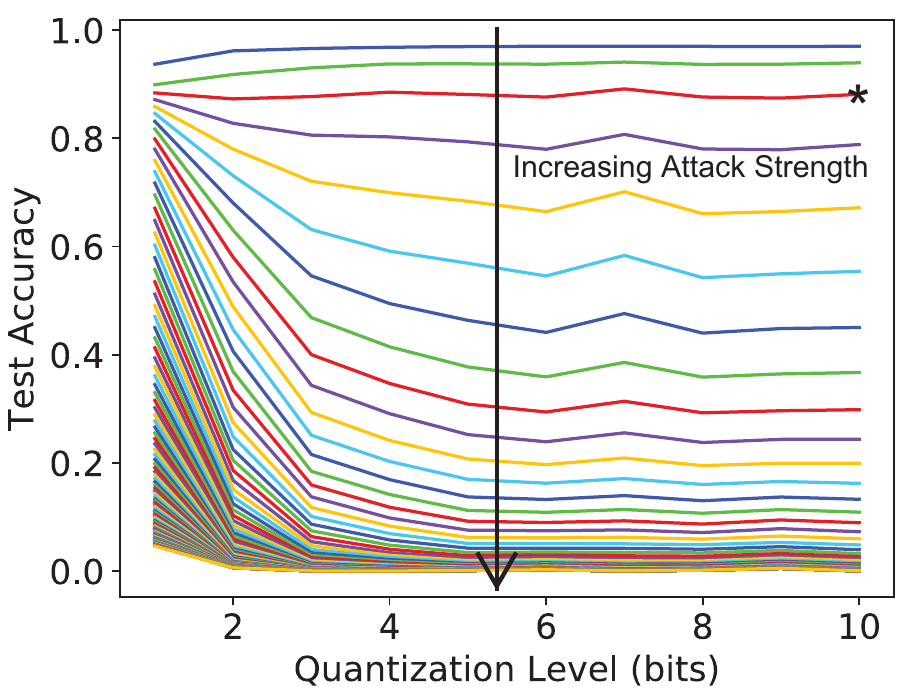}
}
\subfigure[]{
\includegraphics[width=0.3\textwidth]{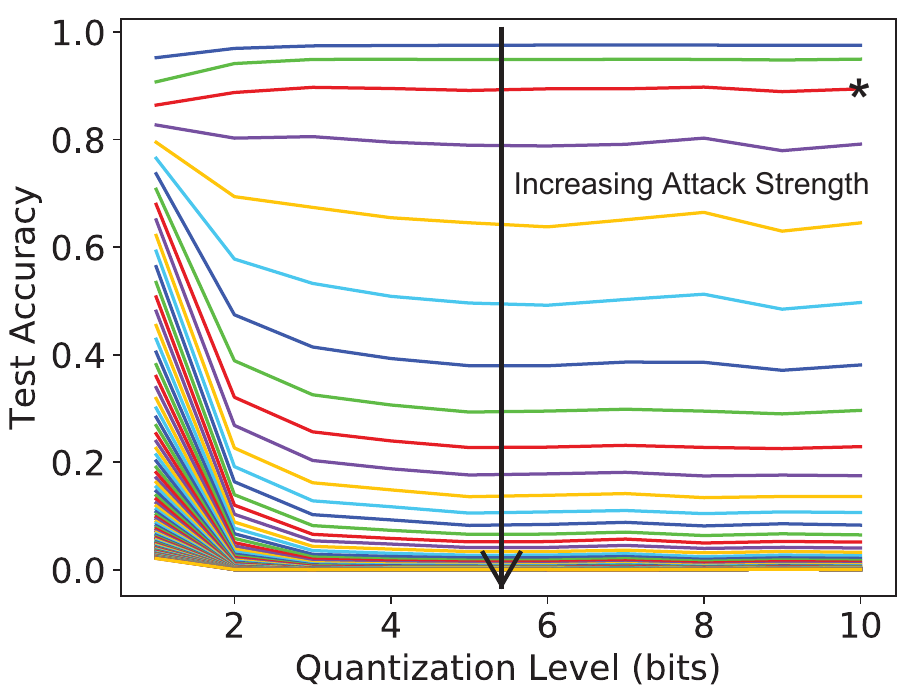}
}
\subfigure[]{
\includegraphics[width=0.3\textwidth]{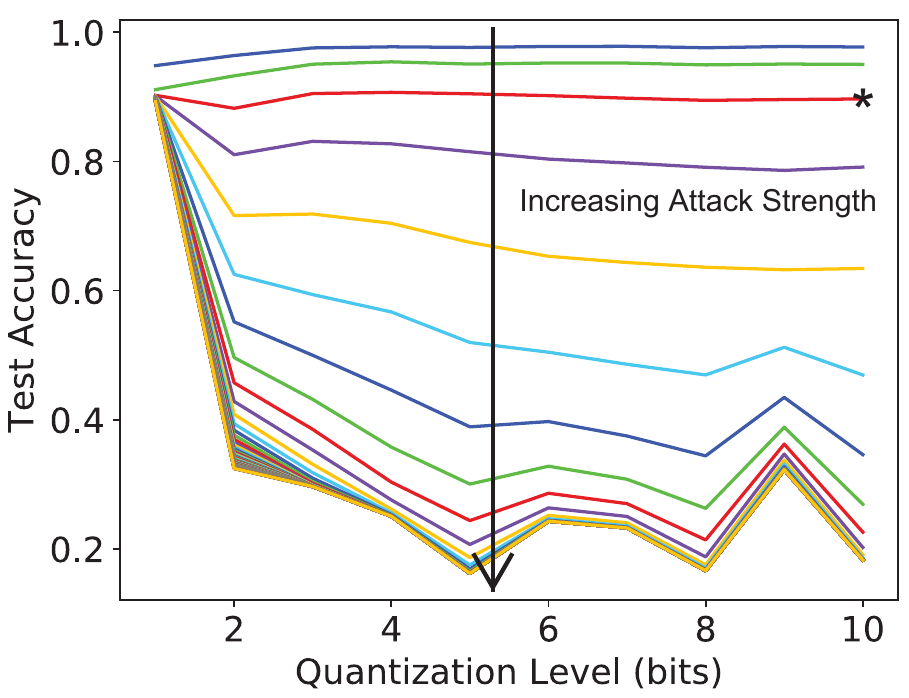}
}
\caption{Test accuracy vs. weight quantization level for different FGSM attack strengths on the MNIST dataset with (a) step activation function, (b) sigmoid activation function and (c) ReLU activation function.  Asterisks denote the curves corresponding to critical attack strengths.}
\label{fig:accvsquantmnist}
\end{figure*}

\section{Results and Analysis}
\label{sec:results}

We performed simulations on two different datasets, MNIST handwritten digits and a 2 spiral classification problem, to study the effect of weight quantization on adversarial robustness against white box FGSM attacks.  For both datasets, we use a multilayer perceptron (MLP) with 100 hidden neurons.  All simulations were performed with Tensorflow and Keras using the Adam optimizer for training with 10\% of the training data held out for validation.  Training proceeded until the accuracy on the validation set reached a maximum value (early stopping).  Besides early stopping, no other regularization (dropout, etc.) methods were employed, and no adversarial defenses were used.  All weight and bias values were limited to the range [-1,1] during training, and quantization was performed using deterministic rounding on the forward pass \cite{guo2018survey}.  All results shown represent the mean of 5 independent simulations with random network initializations.  

\subsection{MNIST Dataset}

The MNIST dataset consists of 28$\times$28 ($n=784$) grayscale images of handwritten digits.  The training set consists of 60,000 images while the test set contains 10,000 images.  For this dataset, our MLP has 784 inputs, 100 hidden neurons, and 10 outputs.  We used a softmax activation at the output with categorical cross entropy loss.  Figure \ref{fig:accvsquantmnist} shows the test accuracy vs. weight quantization level from 1 to 10 bits for three different activation functions in the hidden layer:  step, sigmoid, and ReLU.  Note that we did not see much change in the results beyond 10-bit weights.  In each plot, we also show a sweep of FGSM attack strengths from 0 to 0.5 in 0.005 increments.  For all three activation functions, we observe that smaller bit widths tend to improve robustness to FGSM attacks, with binary weights providing the best resilience.  This is consistent with other studies such as \cite{galloway2017attacking} which evaluated binarized neural networks in an adversarial setting.  Also of interest is the fact that ReLU activation provided the best overall protection, especially at 1-bit quantization.  Potential reasons for this could be gradient masking or the piecewise linear behavior of ReLU ANNs.  However, more analysis is needed to better understand the effect of the activation function on adversarial robustness.

For all of the activation functions, increased weight precision tends to increase the test accuracy when the attack strength is very small.  However, after a certain value of the attack strength, increasing precision causes a decrease in accuracy.  This highlights the two competing effects of quantization/precision.  On one hand, increased precision allows the network to define tighter decision boundaries between classes, which may be necessary when two classes are close together along one or more feature dimensions.  On the other hand, reducing precision helps to increase the average distance between test data and decision boundaries, which increases the amount of adversarial perturbation necessary to cause the input to change classes.  Interestingly, this also means that there must be a \textit{critical attack strength} where the two effects of quantization balance each other and lead to little or no change in accuracy.  We found the critical attack strength by searching for the accuracy vs. quantization level curve that had the smallest variance for the set of attack strengths that we simulated.  For the MNIST dataset the critical attack strength was 0.01.  The curves corresponding to this critical strength are denoted in Figure \ref{fig:accvsquantmnist} with asterisks.  Notice that these curves are approximately flat compared to the others.  Being able to identify this critical attack strength may help give insight into the characteristics of ANNs, datasets, and attack types where quantization can improve adversarial accuracy.

\begin{figure}
\centering
\includegraphics[scale=0.4]{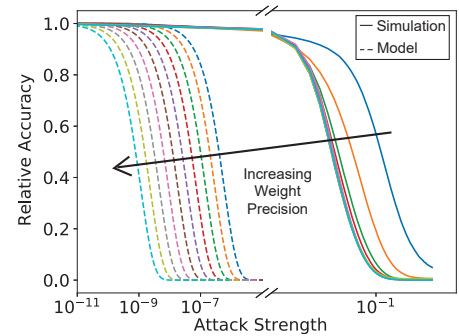}
\caption{Relative accuracy on MNIST vs. attack strength for different levels of weight quantization.}
\label{fig:ramnist}
\end{figure}

We also studied the relative accuracy of the MLP on adversarial MNIST images vs. the attack strength in comparison with our model proposed in (\ref{eqn:ra}).  The lower bound accurately captures the sharp drop in accuracy for small perturbations caused by datapoints that lie very close to or on the decision boundaries.  It also accurately captures the shape of the relative accuracy vs. attack strength.  In general, though, the results show that in practice, we generally do much better than the worst-case lower bound, especially as the attack strength increases.  There are several possible reasons for this.  Notably, our relative accuracy model makes some very conservative assumptions about how close decision boundaries are drawn to datapoints during the training process.  We also assume that all decision regions are the same size and that the datapoints are uniformly distributed within each region, both of which may need to be revised in order to tighten our lower bound model for larger attack strengths.  Note that we have also implicitly used an adversarial defense technique in our simulations—regularization via early stopping, which we have not modeled in our lower bound.  However, we believe this is a good foundation for modeling the worst-case effect of adversarial perturbations on quantized models and warrants further exploration.

\subsection{Spiral Dataset}

\begin{figure}
    \centering
    \includegraphics[scale=0.4]{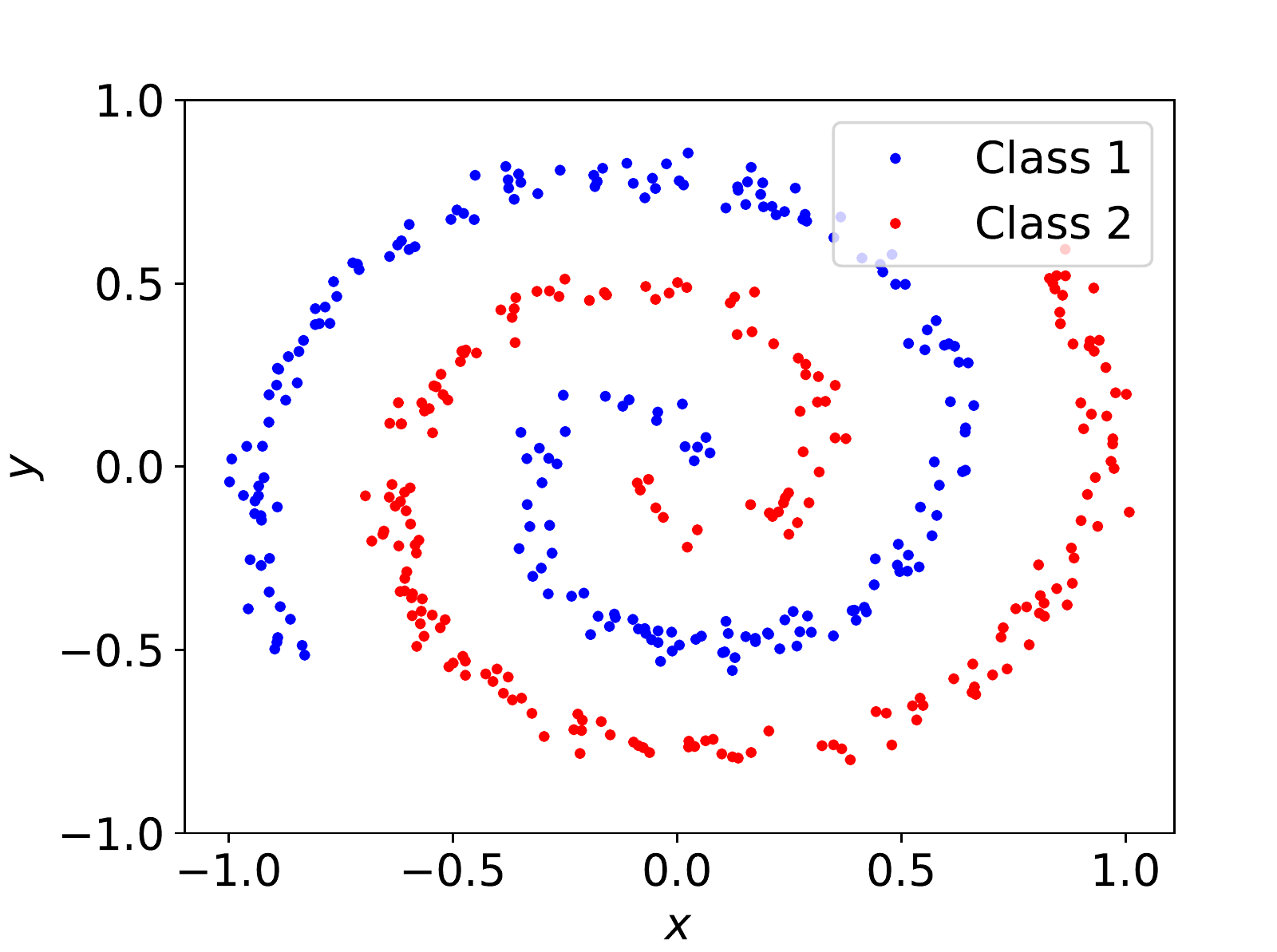}
    \caption{Example of 2 spiral classification problem.}
    \label{fig:spiral}
\end{figure}

\begin{figure*}[!t]
\centering
\subfigure[]{
\includegraphics[width=0.3\textwidth]{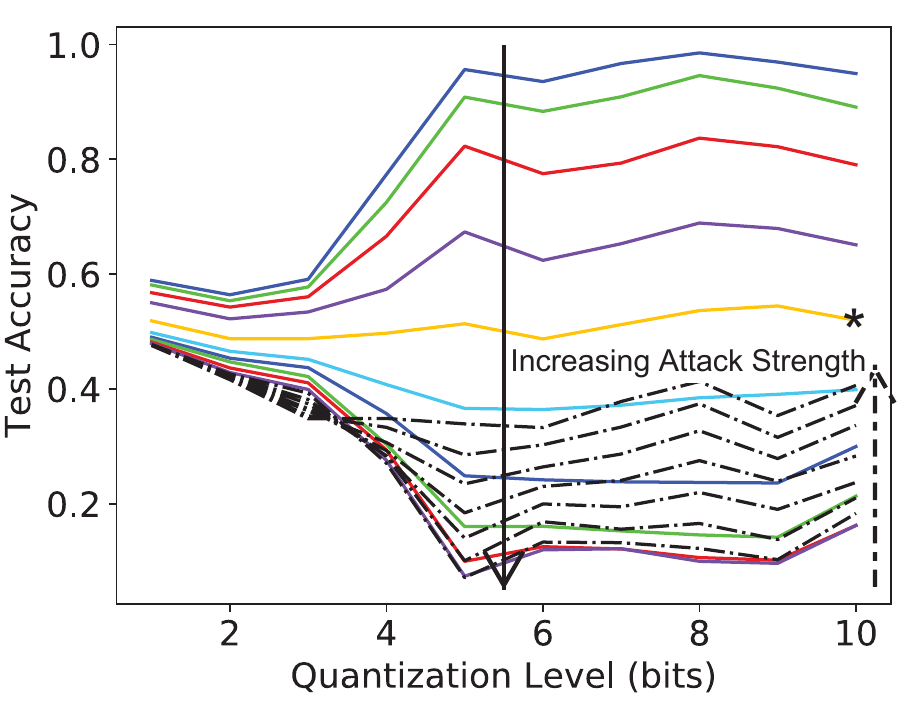}
}
\subfigure[]{
\includegraphics[width=0.3\textwidth]{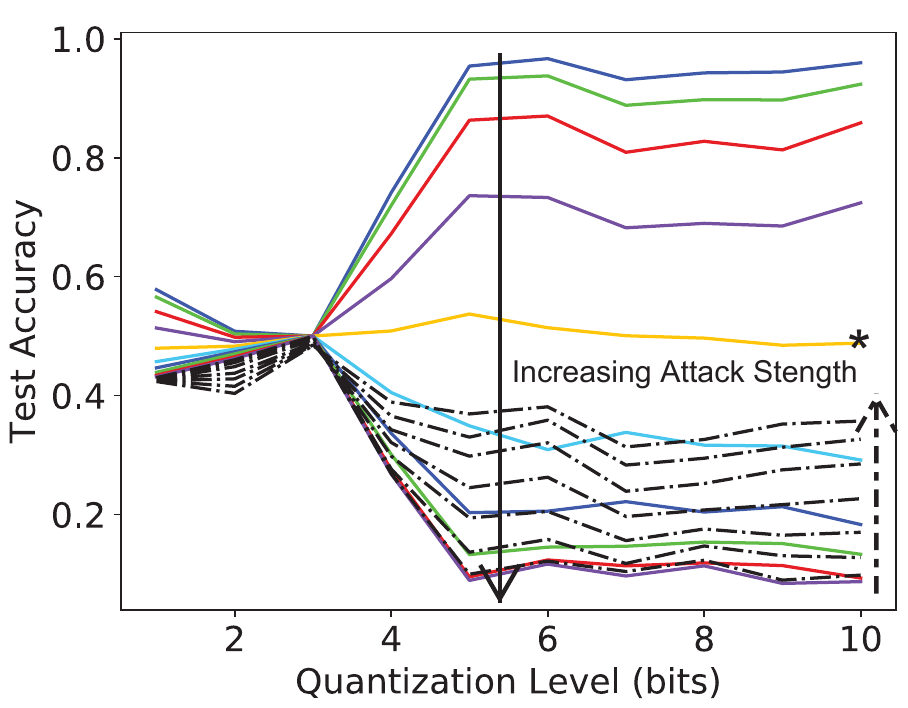}
}
\subfigure[]{
\includegraphics[width=0.3\textwidth]{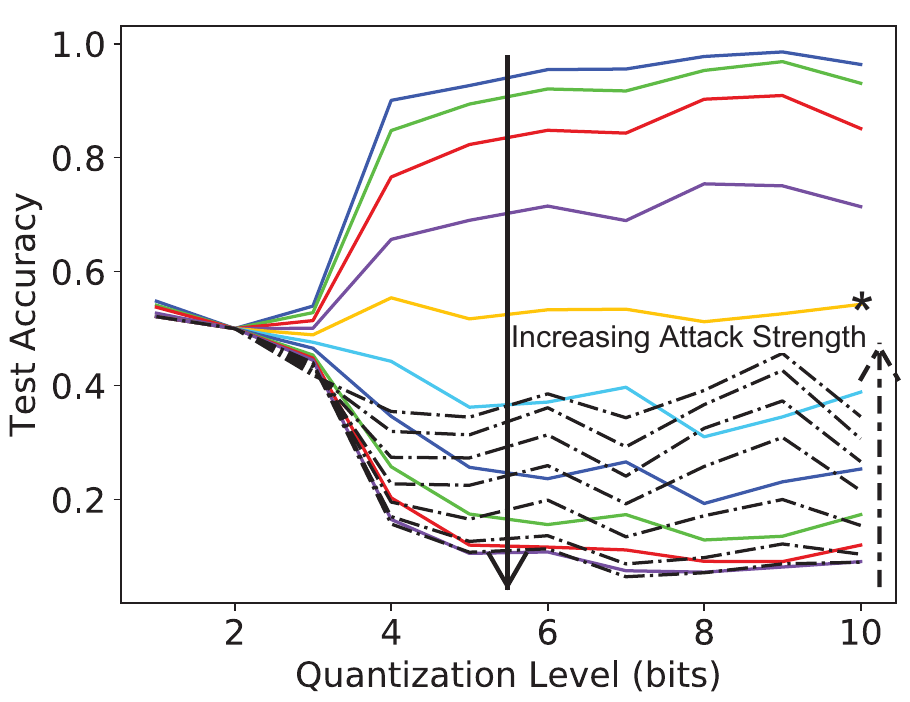}
}
\caption{Test accuracy vs. weight quantization level for different FGSM attack strengths on the spiral dataset with (a) step activation function, (b) sigmoid activation function and (c) ReLU activation function.  Asterisks denote the curves corresponding to critical attack strengths.}
\label{fig:accvsquantspiral}
\end{figure*}

In addition to the relatively high-dimensional MNIST dataset, we also studied the 2 spiral classification problem, which is a low-dimensional (2D) binary classification problem composed of 2 spirals, which were generated with
\begin{equation}
\begin{aligned}
x&=\pm0.1t\mathrm{cos}(t)+\xi_{x}\\
y&=\pm0.1t\mathrm{sin}(t)+\xi_{y}
\end{aligned}
\end{equation}
where $t$ is distributed uniformly between 0 and 100, and $\xi_{x}$ and $\xi_{y}$ are normally-distributed with mean 0 and variance 0.03.  Taking the $+$ value for $x$ and $y$ gives one spiral, while taking the $-$ value gives the other spiral.  Figure \ref{fig:spiral} shows an example.   In this paper, we used 200 points in each spiral for both the training and test sets.  Figure \ref{fig:accvsquantspiral} shows the accuracy vs. weight quantization level across different FGSM attack strengths (same range as in MNIST simulations) and different hidden activation functions with an MLP that has 2 inputs, 100 hidden units, and 1 output.  Since this there is only a single output, we used a sigmoid activation function at the output layer with binary cross entropy loss.  In addition, to achieve high accuracy at high weight precision, we amplified the inputs to the hidden unit activations by 10, effectively scaling all weights to lie in the range $[-10,10]$.  We note a couple of similarities and differences compared to the MNIST results.  First, even at 0 attack strength, low bit width quantization brings the accuracy close to random chance (50\%) across all activation functions.  This is in contrast to the MNIST results, where even 1-bit weight quantization gave clean accuracies above 90\%.  We also observe, as in the MNIST case, that the MLPs employing ReLU activations have the least sensitivity to changes in weight quantization at low bit widths.  The competing effects of quantization that appeared as a bifurcation in the MNIST data can also be seen in the 2 spiral classification results.  The critical attack strength in this case is $\approx$0.125 ($\ell^{2}$ distance of 0.18) across all three activation functions.  This is significantly smaller than the critical distance for MNIST (0.28).

\begin{figure}
\centering
\includegraphics[scale=0.6]{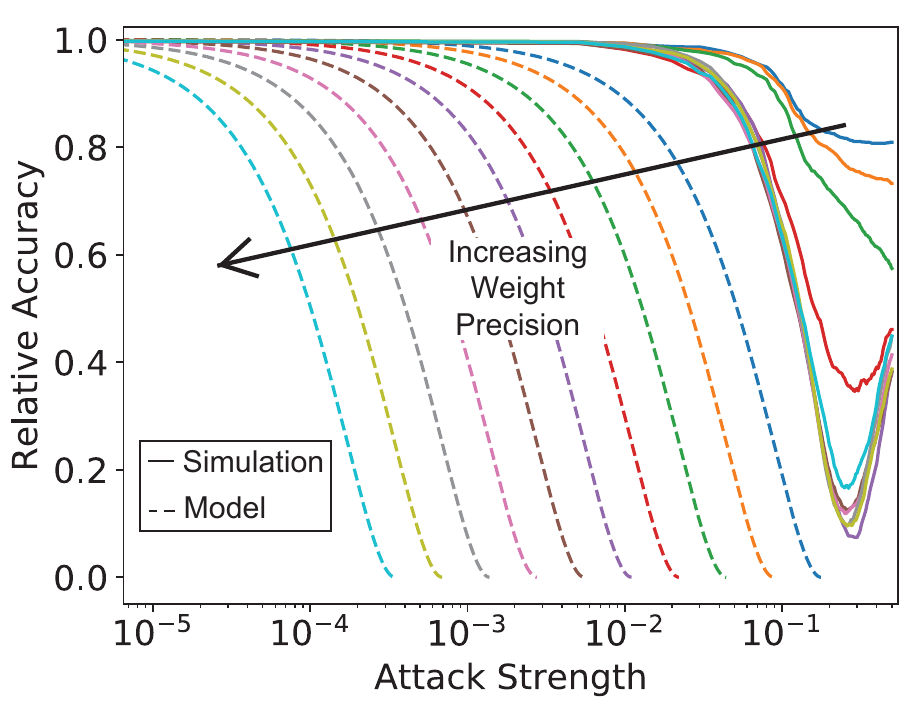}
\caption{Relative accuracy on the 2 spiral problem vs. attack strength for different levels of weight quantization.}
\label{fig:raspiral}
\end{figure}

Another interesting thing to note about the results for the 2 spiral classification problem is that as the attack strength increases, the accuracy drops to $\approx$0\% (indicated by solid colored lines) but then starts to increase again (indicated by dash-dot black lines).  The reason for this can be extracted from Figure \ref{fig:spiral}.  For example, pick a point from Class 1 and imagine moving in the direction of the closest point from Class 2.  Eventually, you will move into a Class 2 region.  However, if you keep moving in the same direction, there is a good chance that you will move back into a Class 1 region.  Figure \ref{fig:raspiral} shows the relative accuracy vs. attack strength for the 2 spiral classification problem along with the lower bound from out model.  In this case, the lower bound is much closer to the the simulation data (by several orders of magnitude) than in the case of the MNIST results because of the lower dimensionality of the input space.

\section{Conclusions and Future Work}
\label{sec:conclusions}

In this paper, we have explored the effect of weight quantization on the adversarial robustness of ANNs.  We showed that there is a critical adversarial attack strength at which quantization has little-to-no effect on accuracy.  For attack strengths less this critical strength, increasing weight precision improves accuracy by enabling more complex decision boundaries.  At attack strengths greater than the critical strength, increasing precision causes a drop in accuracy stemming from decision boundaries being closer to datapoints.  We also compared our results to a novel geometric model of adversarial robustness based on the geometry of the decision region and found that in practice, adversarial robustness is much higher than the worst case predicted by the model.  Targets for future work include the exploration of different ANN topologies, model refinement by considering non-convex decision regions with non-uniform data distributions, and further analysis of the effect of activation function on the performance of weight quantized ANNs. 

\section*{Acknowledgements}

This material is based on research sponsored by the Air Force Research Laboratory under agreement number FA8750-20-2-0503. The U.S. Government is authorized to reproduce and distribute reprints for Governmental purposes notwithstanding any copyright notation hereon.  The views and conclusions contained herein are those of the authors and should not be interpreted as necessarily representing the official policies or endorsements, either expressed or implied, of the Air Force Research Laboratory or the U.S. Government.

\bibliographystyle{ACM-Reference-Format}
\bibliography{refs.bib}

\end{document}